\documentclass[runningheads]{llncs}
\usepackage{graphicx}
\usepackage[colorlinks=true, allcolors=blue]{hyperref}
\usepackage{amsmath,amsfonts,amssymb}
\usepackage{algorithm2e} 
\usepackage{algpseudocode} 
\usepackage{subcaption}
\SetKwInput{KwInput}{Input}                
\SetKwInput{KwOutput}{Output}              
\usepackage[T1]{fontenc} 
\usepackage[normalem]{ulem}

\newcommand{\MKeditConfirmed}[2]{{\color{black}#2}}
\newcommand\MKreview[1]{\textcolor{black}{#1}}
\newcommand{\R}{{\mathbb R}}
\newcommand{\Mesh}{\mathcal{M}}
\newcommand{\Vertex}{\mathcal{V}}
\newcommand{\Face}{\mathcal{F}}
\newcommand{\Template}{\mathcal{T}}
\newcommand{\Texture}{\mathcal{I}}
\newcommand{\Signal}{F}
\newcommand{\Lesion}{\mathcal{L}}
\newcommand{\VF}{\vec{v}}
\newcommand\norm[1]{\Vert#1\Vert}

\DeclareMathOperator*{\argmin}{arg\,min}

\begin{document}
\title{Revisiting Lesion Tracking in 3D Total Body Photography}
%
%
\author{Wei-Lun Huang\inst{1,4} \and
Minghao Xue\inst{1,4} \and
Zhiyou Liu\inst{5} \and
Davood Tashayyod\inst{5} \and
Jun Kang\inst{2}\and
Amir Gandjbakhche\inst{4} \and
Misha Kazhdan \inst{1} \and
Mehran Armand \inst{3}
}
\authorrunning{W. Huang et al.}
%
\institute{Johns Hopkins University, Whiting School of Engineering, Department of Computer Science, Baltimore, MD, USA \and
Johns Hopkins School of Medicine, Department of Dermatology, Baltimore, MD, USA \and
University of Arkansas, Institute for Integrative and Innovative Research, Department of Mechanical Engineering, AR, USA \and
Eunice Kennedy Shriver National Institute of Child Health and Human Development, Bethesda, MD, USA \and
Lumo Imaging, Rockville, MD, USA
}
\maketitle              
\begin{abstract}
Melanoma is the most deadly form of skin cancer. Tracking the evolution of nevi and detecting new lesions across the body is essential for the early detection of melanoma. Despite prior work on longitudinal tracking of skin lesions in 3D total body photography, there are still several challenges, including 1) low accuracy for finding correct lesion pairs across scans, 2) sensitivity to noisy lesion detection, and 3) lack of large-scale datasets with numerous annotated lesion pairs. We propose a framework that takes in a pair of 3D textured meshes, matches lesions in the context of total body photography, and identifies unmatchable lesions. We start by computing correspondence maps bringing the source and target meshes to a template mesh. Using these maps to define source/target signals over the template domain, we construct a flow field aligning the mapped signals. The initial correspondence maps are then refined by advecting forward/backward along the vector field. Finally, lesion assignment is performed using the refined correspondence maps. We propose the first large-scale dataset for skin lesion tracking with 25K lesion pairs across 198 subjects. The proposed method achieves a success rate of 89.9 \% (at 10 mm criterion) for all pairs of annotated lesions and a matching accuracy of 98.2\% for subjects with more than 200 lesions.

\keywords{Total body photography  \and Skin lesion longitudinal tracking \and 3D correspondence.}
\end{abstract}

\section{Introduction}
\label{sec:intro}

Melanoma is the most deadly form of skin cancer. Tracking the evolution of existing nevi and detecting new lesions across the body is essential for the early detection of melanoma \cite{abbasi2004early}. Manual evaluation of skin lesions by a dermatologist is considered the standard of care. However, in patients with numerous skin lesions, this task becomes challenging and is prone to human error.

Total body photography (TBP) captures the entire body of a patient using 2D images \cite{halpern2003total} and/or a 3D mesh \cite{rayner2018clinical,grochulska2022additive}. As such, TBP can be effective for monitoring the evolution of lesions \cite{deinlein2020importance,hornung2021value,ji2021total,guido2021novel,primiero2019evaluation,primiero2024narrative}. From a systematic review, an individual with a large number (>100) of common nevi is usually included in the target population of TBP \cite{hornung2021value}. 

Several prior works have proposed the use of 3D meshes for longitudinal tracking of skin lesions \cite{bogo2014automated,zhao2022skin3d,ahmedt2023monitoring,huang2023skin}. However, they continue to face significant challenges.
First, existing methods produce low-accuracy lesion correspondences, attributed to multiple factors such as the limited positional expression of a lesion, regions with non-isometric deformations, and inconsistent texture across scans. Methods proposed in \cite{zhao2022skin3d,ahmedt2023monitoring,huang2023skin} are constrained by their representation of lesions at the resolution of the mesh vertices, thereby making them heavily dependent on the resolution of the mesh. In extreme cases, adjacent lesions will be mapped to the same vertex, reducing tracking accuracy. In addition, approaches from Zhao \textit{et al.} \cite{zhao2022skin3d} and Ahmedt-Aristizabal \textit{et al.} \cite{ahmedt2023monitoring} rely on a coarse correspondence map between the source and target meshes to identify corresponding lesion pairs. Thus, even with an accurate representation of the lesions themselves, correspondences are necessarily imprecise.
Huang \textit{et al.} \cite{huang2023skin} improve the correspondence accuracy by incorporating both shape and texture information. However, their method is sensitive to inconsistent texture across scans, caused by imperfect scanning (e.g. misalignment between images) or changes in clothing and hairiness.

Independently, noise in lesion detection is unavoidable in practice \cite{zhao2022skin3d}, including both false positives (detections that are not actual lesions) and false negatives (lesions that are undetected). Therefore, matching methods need to not only identify correspondences between of inlier lesions on the source/target (i.e., lesions that are successfully detected in both scans) but also provide the corresponding location on the target/source for lesions that cannot be matched. This additional location information allows physicians to verify whether an unmatchable lesion represents a new growth or is a false positive resulting from noise.

The public dataset introduced by Zhao \textit{et al.} \cite{zhao2022skin3d} represents a pioneering effort in lesion tracking using 3D meshes. However, the size of the dataset is limited, comprising only 10 subjects. Furthermore, on average only 20 lesion pairs are annotated per subject. The sparse annotation of skin lesions makes the lesion-tracking evaluation far from representative of real-world scenarios.


We propose a framework to match lesions in the context of the TBP using 3D textured meshes while providing locations for unmatchable lesions. To achieve this, we compute accurate correspondence maps relying on the signal represented by the TBP images, in addition to the geometry of the meshes themselves. Using the geometry of the meshes, we first compute coarse correspondence maps taking the source and target meshes to a template mesh. Then, using  the lesion/texture signals, we solve for a flow field on the template mesh which is used to refine the correspondence maps. Finally, lesion assignment is performed using the refined correspondence maps. We also extend the annotations on the 3DBodyTex dataset \cite{saint20183dbodytex} to a 25K lesion pairs dataset for skin lesion tracking. To the best of our knowledge, we are the first to release a dataset for skin lesion tracking dataset at this scale.

Overall, we make three main contributions:
\begin{itemize}
    \item We propose a novel framework for lesion tracking that automatically matches inlier lesions while providing locations for unmatchable lesions in the context of TBP using 3D textured mesh.
    \item We extend the 3DBodyTex dataset by annotating 25K lesion pairs over 198 subjects for skin lesion tracking in 3D TBP.       
    \item We validate that the proposed framework outperforms the state-of-the-art methods in both the matching accuracy and the accuracy of the correspondence maps. The framework is also more robust to 1) inconsistency between source and target texture, 2) non-isometric deformations, and 3) errors in lesion detection.
    
\end{itemize}


\section{Related work}
\label{sec:literature}

\subsection{Shape correspondence for humans}
\label{ss:shape correspondence}
Shape correspondence between non-rigid surfaces represented as triangle meshes has been an active research topic in computer vision and computer graphics \cite{van2011survey,sahilliouglu2020recent,deng2022survey}. The shape correspondence problem for triangle meshes is finding a set of corresponding points between two meshes. For human shapes, priors of the human body are commonly applied, such as near-isometric deformation and local rigidity for limbs \cite{zuffi2015stitched,wang2021locally,bhatnagar2020combining,alldieck2021imghum}.

\paragraph{Template-based} A line of research relies on a template model, such as SMPL \cite{loper2023smpl}, for establishing correspondences across shapes \cite{huang2020dense,groueix20183d,bhatnagar2020loopreg}. These methods usually rely on an initial estimation of the pose (body joint positions and orientation) mapping the template to the input. Groueix \textit{et al.} \cite{groueix20183d} propose to deform a template mesh to various body poses and shapes with auto-encoder frameworks. Bhatnagar \textit{et al.} \cite{bhatnagar2020loopreg} use self-supervised learning to register scans of humans to a common 3D human model.

\paragraph{Canonical embedding} Another common shape correspondence method maps vertices into a pose-invariant feature space, where correspondences between the input and template geometries are more easily established \cite{ovsjanikov2012functional,litany2017deep,marin2020farm,donati2020deep}. In this category, several shape descriptors have been proposed, from traditional hand-crafted descriptors \cite{sun2009concise,salti2014shot,mitchel2021echo} to deep-learning-based descriptors \cite{wei2016dense,sharp2022diffusionnet,mitchel2021field,kim2021deep}. Furthermore, functional map \cite{ovsjanikov2012functional} are commonly used for robust regularization.

However, for matching skin lesions, the correspondence map relying on the geometry is not sufficiently accurate. Therefore, using the coarse correspondence map may fail to pair up lesions, particularly if the subject undergoes non-isometric deformation from scan to scan with numerous lesions in close vicinity. We propose leveraging additional signals on the mesh to refine the correspondence map for more accurate matching.
\subsection{Graph matching}
\label{subsection:graph matching}
\MKreview{Given a set of lesions detected in a mesh, we can construct a graph in which a node corresponds to a single lesion and an edge is a connection between a pair of lesions. The node attribute is the position of a lesion, and the edge attribute is the geodesic distance between a pair of lesions. Then, the problem of matching source and target lesions can be formulated as a (partial) graph-matching problem that maximizes the node-to-node and edge-to-edge affinity of the two graphs. Two-graph matching can be modeled as the quadratic assignment problem (QAP), and is known to be NP-hard.} 

Traditional approaches \cite{leordeanu2009integer,enqvist2009optimal,cho2010reweighted} aim to match graphs by maximizing quadratic objective functions. While effective for simple cases, these methods often struggle with complex graph structures. Some proposed approaches utilize relaxation strategies in graph matching to mitigate the hard combinatorial problem \cite{leordeanu2005spectral,kang2013fast,schellewald2005probabilistic}. More recent methods explore hyper-graph matching represented by a tensor to encode the higher-order information, offering increased expressiveness but at the cost of higher computational complexity \cite{duchenne2011tensor,park2013fast,yan2015discrete}. 

Learning-based methods have been shown to improve matching accuracy \cite{wang2021neural,rolinek2020deep,liao2021hypergraph}. \MKreview{Wang \textit{et al.} \cite{wang2021neural} presents a QAP network to solve the matching problem as a vertex classification task over the association graph whose nodes represent candidate correspondences between the two graphs and edge weights are induced by the affinity matrix built with the two graphs.} Liao \textit{et al.} \cite{liao2021hypergraph} convert the problem of hypergraph matching into a node classification problem and develop a hypergraph neural network. Despite their promise, these methods often require exhaustively annotated datasets for training and struggle to generalize across different domains or datasets. Meanwhile, to address real-world scenarios involving noisy or incomplete data, some techniques are developed for partial graph matching and soft assignment \cite{wang2023deep,fu2021robust}.
 
Despite the advances in graph matching, the solution itself (unlike a correspondence map) does not provide location information for unmatched lesions, an issues that is critical in clinical settings. Starting from coarse correspondence maps, we propose a framework that performs lesion assignment and maps lesions onto a template mesh that allows clinical verification.  

\subsection{Skin lesion tracking in total body photography}
\label{subsection:skin lesion}
Several works have been proposed for tackling the skin lesion tracking problem over the full body \cite{korotkov2018improved,korotkov2014new,strkakowska2022skin,strzelecki2021skin}. Korotkov \textit{et al.} \cite{korotkov2014new} designed a TBP system with 21 high-resolution cameras and a turntable to track lesions. However, their method assumes the patient poses are the same across visits and heavily rely on calibrated camera poses for finding lesion correspondence. The work of Korotkov \textit{et al.} \cite{korotkov2018improved} improved the earlier method but does not extend to multiple visits. Strzelecki \textit{et al.} \cite{strzelecki2021skin} developed a TBP system with one digital camera rotating and moving vertically around the subject. The camera captures 32 images for lesion detection and lesion matching \cite{strkakowska2022skin} based on feature matching and triangulation. However, their method fails when the skin surface is inclined at an angle deviating significantly from 90$^\circ$ with respect to the camera viewing direction. Overall, these methods are limited to a controlled environment and sensitive to camera perspectives and changes in body poses \cite{huang2023skin}.

Recently, the concept of finding lesion correspondence using a 3D representation of the human body has been explored in \cite{bogo2014automated,zhao2022skin3d,ahmedt2023monitoring,huang2023skin}. 
Zhao \textit{et al.} \cite{zhao2022skin3d}, Bogo \textit{et al.} \cite{bogo2014automated}, and Ahmedt-Aristizabal \textit{et al.} \cite{ahmedt2023monitoring} proposed to use a template mesh and rely on anatomical position defined on the template mesh for lesion matching. However, using the template-based correspondence map is insufficient to accurately match lesions. Furthermore, these methods have limited positional expression for lesions since they ``snap'' the location of a lesion to the nearest vertex.
Huang \textit{et al.} \cite{huang2023skin} proposed to improve the lesion correspondence localization accuracy using landmark-based correspondences refined by texture information. However, their method requires manual annotation of landmarks, is limited to the resolution of the mesh, and is sensitive to \MKreview{inconsistent texture between scans due to scanning artifacts.} In this paper, we represent lesion positions using barycentric coordinates within a triangle, making our approach less sensitive to the resolution of the mesh and allowing us to achieve a higher matching accuracy. \MKreview{Additionally, we utilize lesion signals that are agnostic to the textured mesh assuming lesions are provided separately.} 

\section{Methods}

\subsection{Problem Formulation}
Given a template mesh $\Mesh_{\Template}$, source and target meshes $\Mesh_0$, $\Mesh_1$, and two sets of detected lesions $X_0 \subset \Mesh_0$, $X_1 \subset \Mesh_1$, we would like to find correspondence maps $\phi_0^{\Template}: \Mesh_0 \rightarrow \Mesh_{\Template}$ and $\phi_1^{\Template}: \Mesh_1 \rightarrow \Mesh_{\Template}$, and a matching matrix $\pi=\{0,1\}^{(|X_0| + 1) \times (|X_1| + 1)}$ \MKreview{minimizing an energy of consisting of two terms:
\begin{equation}
\label{eqn:energy all}
    E_{X_0, X_1}(\phi_0^{\Template}, \phi_1^{\Template}, \pi) = \sum_{X_0,X_1}E_{\text{DistanceProximity}}(\phi_0^{\Template}, \phi_1^{\Template}, \pi) + E_{\text{Stochasticity}}(\pi) \ .
\end{equation}
That is, we want a pair of corresponding source and target lesions to be close to each other while encouraging the correspondence matrix to be doubly stochastic.}
By adding a dummy lesion to each of the lesion sets in $\pi$, we allow the matching function to account for unmatchable lesions. Specifically, assuming a lesion in the source scan can be matched to at most one lesion in the target scan and vice versa, we also enforce:
\begin{align}
\sum_{j=0}^{|X_1|} \pi_{i,j} &= 1, \quad \forall i=1,\ldots,|X_0| \label{eq:1} \\
\sum_{i=0}^{|X_0|} \pi_{i,j} &= 1, \quad \forall j=1,\ldots,|X_1| \label{eq:2}
\end{align}
(where $\pi_{i,0}=1$ indicates a match between the $i$-th lesion on the source and the target's dummy lesion). Since the dummy lesions can be matched multiple times, the sums involving the dummy lesions can be greater than 1. 

\subsection{Coarse Correspondence Map}
\label{ss:coarse correspondence}

\subsubsection{Template-based coarse correspondence}
We start by constructing a coarse correspondence map between the source/target and a template mesh. 
We follow the approach from Marin \textit{et al.} \cite{marin24nicp} to acquire a deformed template mesh registered to the source/target mesh that allows us to construct the correspondence map.
\MKreview{Given an input mesh, they propose a localized neural fields network in which a neural field is dedicated to a local region of body shape to predict the vertex displacement of the template mesh (SMPL \cite{loper2023smpl} model). The parameters of the neural field are then refined using Iterative Closest Point \cite{besl1992method} through backpropagation. Then, the updated neural field is utilized to register the SMPL model to the input, followed by a refinement that optimizes Chamfer distance.
We denote the method SMPL-NICP. Let $\Mesh_{\Template}$ be the template mesh, for an input mesh $\Mesh_i, \ i=\{0,1\}$, the output from SMPL-NICP is a deformed template mesh (i.e. with the same topology as the original template) whose geometry is registered to that of $\Mesh_i$.
}

\begin{figure}[]
   \begin{center}
   \includegraphics[width=11cm]{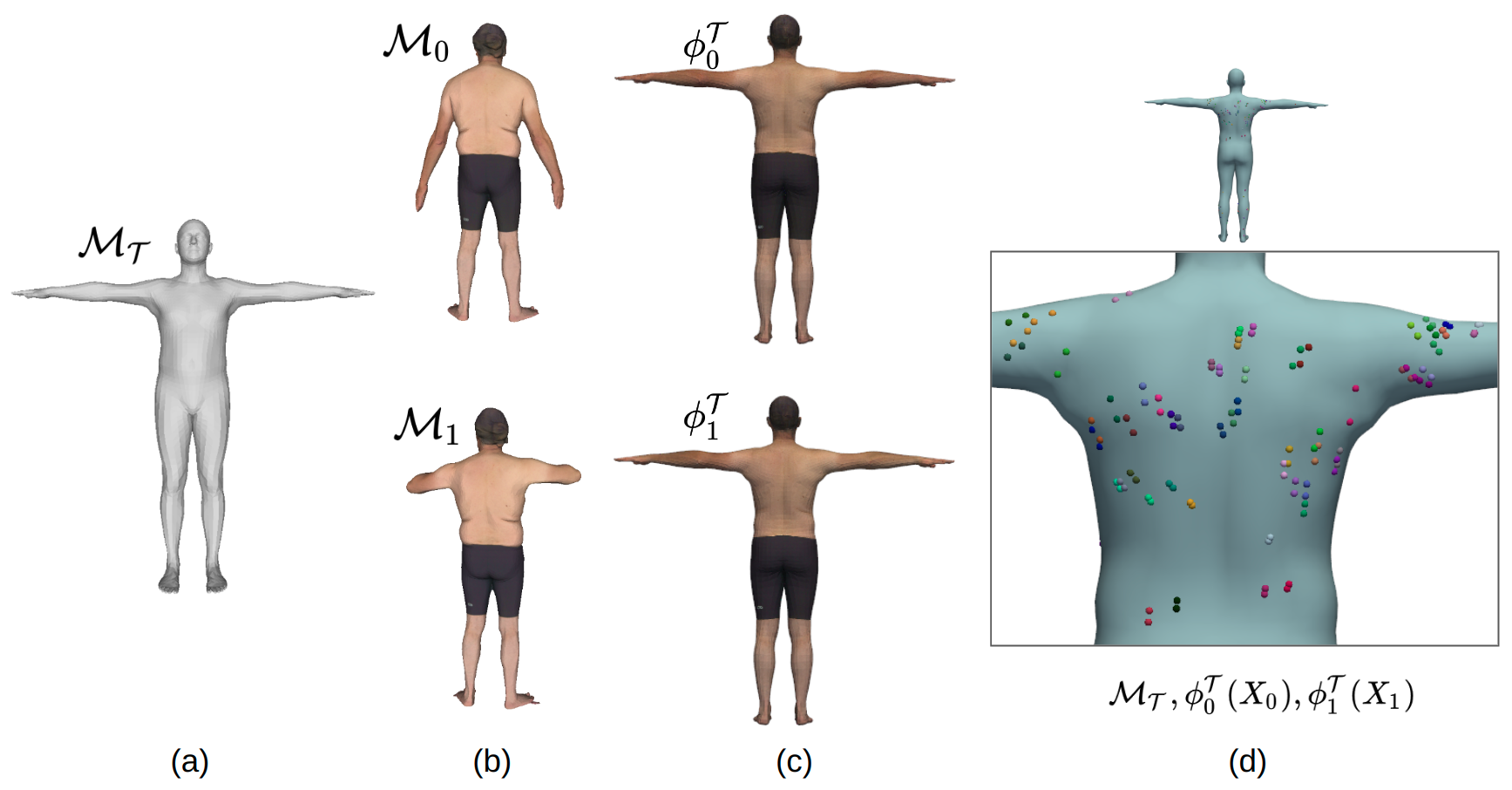}
	\end{center}
   \caption[example] 
    { \label{fig:coarse phi}
Visualization of the source and target lesions mapped to a template mesh by registering the template mesh to the source and target meshes. (a) shows the template mesh $\Mesh_\Template$. (b) shows the source mesh ($\Mesh_0$) and the target mesh ($\Mesh_1$). (c) shows the correspondence maps from the source/target to the template. (d) shows the source and the target lesions mapped to the template mesh. Lesions in correspondences are visualized in the same color.} 
\end{figure}

\MKreview{We define a correspondence map $\phi_i^{\Template}: \Mesh_i \rightarrow \Mesh_{\Template}$, by first deforming the template mesh to $\Mesh_i$ and then finding, for every point $p \in \Mesh_i$, the nearest surface point on the deformed template.}
\MKreview{Similarly, we construct a correspondence map $\phi_{\Template}^i: \Mesh_{\Template} \rightarrow \Mesh_i$ by finding the closest surface point on the input mesh $\Mesh_i$ for each point on the deformed template mesh.
}
We note that $\phi_i^{\Template}$ and $ \phi_{\Template}^i$ are not inverses of each other since two different points on the source/target can have the same closest point on the deformed template. Fig.~\ref{fig:coarse phi} illustrates the template mesh in (a), the source and the target meshes in (b), the correspondence maps from the source/target to the template in (c), and the source and the target lesions mapped to the template mesh in (d). 

\subsubsection{Surface point correspondence map}
We allow lesions to be located anywhere on the surface on the mesh (i.e. not restricted to vertex positions, as in previous work \cite{zhao2022skin3d,ahmedt2023monitoring,huang2023skin}). To this end, we use barycentric coordinates to encode a point on the mesh: $p\in\Mesh\leftrightarrow(t^p, \{\alpha^p,\beta^p,\gamma^p\})$ where $t^p$ indexes the triangle containing $p$ and $\{\alpha^p,\beta^p,\gamma^p\}$ are the barycentric coordinates of $p$ inside the triangle ($0\leq\alpha^p,\beta^p,\gamma^p\leq1$ and $\alpha^p+\beta^p+\gamma^p=1$).

Using this encoding, we represent mesh correspondences as vertex-to-surface-point maps, taking the vertices on one mesh to points on the second mesh: $\Phi_i^j: \Vertex_i \rightarrow \Mesh_j$ is represented by a $\R^{|\Vertex_i| \times 4}$ matrix. The $l^{\text{th}}$ row in $\Phi_i^j$ maps the $l^{\text{th}}$ vertex $v_i^l \in \Vertex_i$ to a point in $\Mesh_j$ in the barycentric encoding.

Given a vertex-to-surface-point correspondence map $\Phi_i^j: \Vertex_i \rightarrow \Mesh_j$, we use the barycentric encoding to extend it to 
a surface-point-to-surface-point correspondence map $\phi_i^j: \Mesh_i \rightarrow \Mesh_j$. Concretely, to find the correspondence of a surface point $p \in \Mesh_i$ to the mesh $\Mesh_j$, we map the three vertices of the triangle containing the point $p$ onto the mesh $\Mesh_j$, interpolate the positions of the imaged vertices using the barycentric coordinate of $p$, and then find the point on $\Mesh_j$ closest to the interpolant.
Formally, for a point $p\leftrightarrow(t^p,\{\alpha^p,\beta^p,\gamma^p\}) \in \Mesh_i$ with $\Face_i(t^p)=(v_0^p,v_1^p,v_2^p)$ representing the triangle containing $p$, we have:
\begin{equation}
    \phi_i^j(p) = \argmin_{q \in \Mesh_j}
    || \alpha^p \cdot \Phi_i^j(v_0^p) + \beta^p \cdot \Phi_i^j(v_1^p) + \gamma^p \cdot \Phi_i^j(v_2^p) - q || \ .
\end{equation}

\subsection{Vector-field-based Refinement}
\label{ss:vector field}
The template-based correspondence maps are coarse for two reasons. First, when fitting a template mesh to source/target scan, non-isometric deformation is present at locations near body joints and locations of soft tissues. Second, misalignment between the deformed template mesh and the input mesh occurs if the body pose of the input mesh is far from the canonical ``T'' pose. Since the coarse correspondence map relies on the nearest point on the registered template mesh to the query point, such a misalignment degrades the accuracy of the mapping. Consequently, a pair of corresponding points in the source and the target will not map to the same position on the template mesh. To refine the correspondence map, we transfer the texture and lesion signals of the source/target to the template mesh using the source/target-to-template correspondences. We then construct a vector field on the template mesh that aligns the transferred signals. 

\subsubsection{Signal construction on template mesh}
\label{ss:signal}
Let $\Signal_i: \Mesh_i \rightarrow \R$ be a signal on mesh $\Mesh_i$, we transfer the signal to the template mesh using the correspondence map, to define a signal on the template $\Signal_i^\Template\equiv\Signal_i\circ\phi^i_{\Template}:\Mesh_{\Template}\rightarrow\R$. We consider two types of input signals: 
\paragraph{Texture signal}
We construct a triplet of color signals on the template mesh, using the colors in the texture map acquired by the TBP, $\Texture_0^c,\Texture_1^c:\Mesh_{\Template}\rightarrow\R$, with $c\in\{R,G,B\}$.

\paragraph{Lesion signal}
We construct lesion signals on the template mesh using lesion signals defined on the source/target meshes, $\Lesion_0, \Lesion_1:\Mesh_{\Template}\rightarrow\R$. The source/target lesion signals represent the likelihood of a surface point being a lesion. To create the lesion signal, we diffuse a sum of delta functions centered at the lesion positions and normalize the signal across the surface with the maximum signal value to create a scalar-per-vertex signal.

\subsubsection{Surface optical flow}
\label{ss:surface optical flow}
\begin{figure}[]
   \begin{center}
   \includegraphics[width=\textwidth]{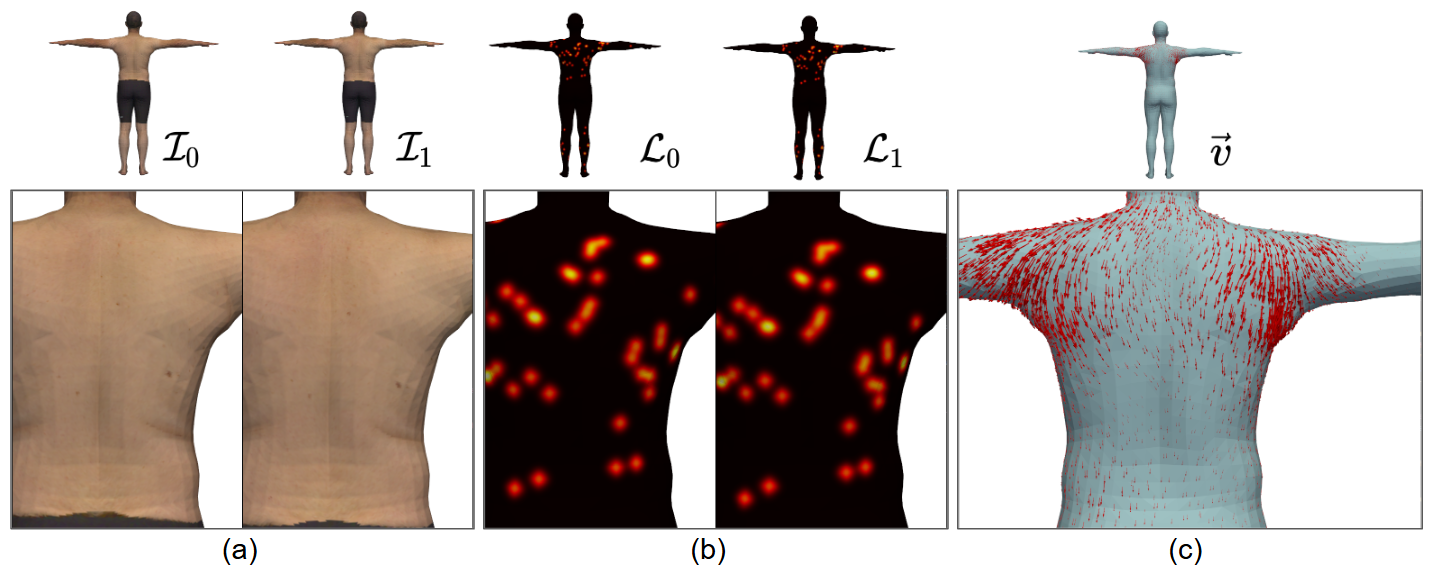}
	\end{center}
   \caption[example] 
    { \label{fig:vector field}
    Visualization of the signals on a template mesh and the vector field. (a) shows the source and the target texture signals. (b) shows the source and the target lesion signals. (c) visualizes a solved vector field that explains the difference between the source and the target signals while being smooth and small.} 
\end{figure}

We are given a template mesh $\Mesh_\Template$, source/target texture signals $\Texture_0^c,\Texture_1^c$, and source/target lesion signals $\Lesion_0,\Lesion_1$.
Our goal is to define a tangent vector field $\VF$ on the template mesh such that advection along the field best aligns the source and target signals. To this end we leverage the approach of Prada \textit{et al.} \cite{prada2016motion}, defining the flow field $\VF$ as the minimizer of the energy:
\begin{equation}
\label{eqn:vf energy}
\begin{aligned}
    E(\VF) = {}&\underbrace{w_{\Texture} \cdot \sum_{\substack{i\in\{0,1\}\\c\in\{R,G,B\}}} \int_{\Mesh_\Template}\Bigl( \langle \nabla \Texture_i^c , \VF \rangle - (\Texture_0^c-\Texture_1)\Bigr)^2\,dp}_{\text{texture fitting}}\\ 
    {}+{}&\underbrace{w_{\Lesion} \cdot \sum_{i=0}^1 \int_{\Mesh_\Template}\Bigl( \langle \nabla \Lesion_i , \VF \rangle - (\Lesion_0-\Lesion_1) \Bigr)^2\,dp}_{\text{lesion fitting}}\\ 
    {}+{}&\underbrace{\epsilon \cdot \int_{\Mesh_\Template}\norm{\nabla\VF(p)}^2\,dp}_{\text{smoothness}} + \underbrace{\varepsilon \cdot \int_{\Mesh_\Template}\norm{\VF(p)}^2\,dp}_{\text{size}}
\end{aligned}
\end{equation}
with the first and the second terms penalizing the failure of the vector field to explain the difference in the texture signal and the lesion signal respectively, the third term encouraging the smoothness of the flow, and the fourth term regularizing the norm of the flow to respect the initial correspondence map.
We follow the approach proposed by Prada \textit{et al.}, solving for the flow field $\VF$ hierarchically. Please refer to \cite{prada2016motion} for more details. Fig.~\ref{fig:vector field} visualizes the source and the target (a) texture and (b) lesion signals transferred to the template mesh. Fig.~\ref{fig:vector field} (c) shows the vector field obtained by minimizing Eqn.\ref{eqn:vf energy}.

\subsubsection{Update of correspondence map}
With the vector field $\VF$ defined on $\Mesh_\Template$, we update the correspondence map $\phi_0^{\Template}$ and $\phi_1^{\Template}$ by advecting the positions of correspondence forward and backward along the vector field halfway, separately.
Formally, we have:
\begin{equation}
    \phi_{0}^{\Template}(p) \leftarrow \exp_{\phi_{0}^{\Template}(p)}
    \frac{\VF\bigl(\phi_{0}^{\Template}(p)\bigr)}{2},\quad\forall p \in \Mesh_0
\end{equation}
and
\begin{equation}
    \phi_{1}^{\Template}(p) \leftarrow \exp_{\phi_{1}^{\Template}(p)}
    \frac{-\VF\bigl(\phi_{1}^{\Template}(p)\bigr)}{2},\quad\forall p \in \Mesh_1
\end{equation}
with $\exp_p:T_p\Mesh_\Template\rightarrow\Mesh_\Template$ the exponential map taking vectors in the tangent space at $p\in\Mesh_\Template$ to positions on $\Mesh_\Template$.


\subsection{Lesion Assignment}
Given source/target correspondence maps $\phi_i^{\Template}:\Mesh_i\rightarrow\Mesh_\Template$, we expect lesions, $x_0\in X_0$ and $x_1\in X_1$ to be in correspondence if the  geodesic distance between $\phi_0^\Template(x_0)$ and $\phi_1^\Template(x_1)$ is small. Conversely, we expect $x_0\in X_0$ (resp. $x_1\in X_1$) to be unmatched if the geodesic distance from $\phi_0^\Template(x_0)$ to $\phi_1^\Template(x_1)$ for all $x_1\in X_1$ (resp. from $\phi_1^\Template(x_1)$ to $\phi_0^\Template(x_0)$ for all $x_0\in X_0$) is large. We formalize these observations, expressing the assignment matrix $\pi\in\{0,1\}^{(|X_0| + 1)\times (|X_1| + 1)}$ as the minimizer of the energy:
\begin{multline}
\label{eqn:matching energy}
    E_{X_0, X_1}(\pi) = 
    \alpha \sum_{x_0 \in X_0, x_1 \in X_1} \pi(x_0, x_1) \cdot D_{\Template}(\phi_0^{\Template}(x_0), \phi_1^{\Template}(x_1)) \ +\\
    \beta \bigl[ \sum_{x_0 \in X_0} \pi(x_0, x_1^{|X_1|+1}) + \sum_{x_1 \in X_1} \pi(x_0^{|X_0|+1}, x_1)\bigr]
\end{multline}
where $D_{\Template}: \Mesh_{\Template} \times \Mesh_{\Template} \rightarrow \R^{\geq0}$ is the geodesic distance function on $\Mesh_{\Template}$. The assignment problem can be optimized through the Kuhn–Munkres algorithms \cite{munkres1957algorithms}. We follow the implementation in Pygmtools \cite{wang2024pygm} to solve the minimizer to Eqn.~\ref{eqn:matching energy}.

\section{Evaluation}


\subsection{Dataset}
\label{ss:dataset}
We extend the 3DBodyTex dataset \cite{saint20183dbodytex,saint2019bodyfitr} by annotating lesion correspondence in every pair of meshes. Following the suggestion from a medical expert reported in \cite{useini2024automatized}, the labeling process is inclusive for anything that could be potentially considered a skin lesion (e.g. including freckles). The dataset is labeled by four experienced annotators using the point list picking function in CloudCompare \cite{cloudcompare}. It takes an annotator approximately 15 minutes to label one subject with 100 lesion pairs in two poses (meshes). We excluded subjects when fewer than 10 lesions were found.

The average number of annotated lesions is 129.6 ($\sigma=88.4$) across 198 subjects, totaling 25,666 lesions. We define the density of the annotated lesions as the number of neighboring lesions within a geodesic distance of 100 mm. Across all subjects, the average density is 8.5 with $\sigma=6.1$. In addition, the lesions are distributed with 12,846 lesions on the trunk, 3761 lesions on the upper right limb, 3617 lesions on the upper left limb, 2139 lesions on the lower left limb, 2291 lesions on the lower left limb, and 1012 lesions on the head. Overall, numerous skin lesion pairs are annotated with diversity in body shapes, sizes, poses, and anatomical variations. Therefore, the proposed dataset is suitable for the evaluation of skin lesion tracking that approximates real-world scenarios for TBP. The distribution of the lesion annotations can be found in the supplement.

We define a ``challenging-pose'' subset that consists of 6 subjects in which one of the two poses is challenging. This subset is separated from the entire dataset and evaluated individually. Furthermore, since lesion tracking is most valuable for patients with numerous and dense lesions, we define another subset of 35 subjects as a ``numerous-lesions'' set in which subjects are annotated with more than 200 lesions.

\subsection{Evaluation of Correspondence Map}
\label{ss:evaluation phi}
To evaluate the quality of the established correspondence maps, we measure the geodesic distance between a pair of source and target lesions mapped on the template mesh. We report 1) the average geodesic distance across all the annotated lesion pairs ($D_{\text{LP}}$) and 2) the subject-wise geodesic distance ($D_{\text{SW}}$) as the average geodesic distance of the annotated lesion pairs for the individual subject, and then aggregated across all the subjects. To interpret the geodesic distance between a pair of source and target lesions in a clinical application, a pair of lesions is successfully mapped if the distance between them is less than a threshold criterion. Using the threshold criterion of 10 mm (as in \cite{huang2023skin}), we measure the success rate for each subject as the percentage of the correctly mapped source and target skin lesions over the total number of annotated skin lesion pairs. We report the subject-wise success rate computed on a pair of meshes (for one subject) and averaged across paired meshes.

We compare our method to two baseline methods that rely solely on geometry to provide correspondence maps, SMPL-NICP \cite{marin24nicp} and DiffusionNet \cite{sharp2022diffusionnet}.
We note that DiffusionNet \cite{sharp2022diffusionnet} is used as a feature extractor to transform each vertex into a higher-order embedding. Therefore, shape correspondence from DiffusionNet \cite{sharp2022diffusionnet} belongs to the category of canonical embedding in shape correspondence (described in \S\ref{ss:shape correspondence}), relying on functional maps to compute correspondence between shapes assuming descriptor preservation over the underlying meshes.
Table~\ref{tab:phi comparison} shows the comparison of the established correspondence maps. For both $D_{\text{LP}}$ and $D_{\text{SW}}$, the correspondence maps from our method are substantially more accurate with a smaller variance compared to the baseline methods. As a result, the proposed method achieves a success rate of 89.9\% ($\sigma=4.7\%$) at the 10-mm criterion, significantly higher than baseline methods. We note that the reported success rate from \cite{huang2023skin} is only 57\% ($\sigma=14\%$), computed over 10 subjects and totaling around 200 lesions within our dataset.

\begin{figure}[]
   \begin{center}
   \includegraphics[width=\textwidth]{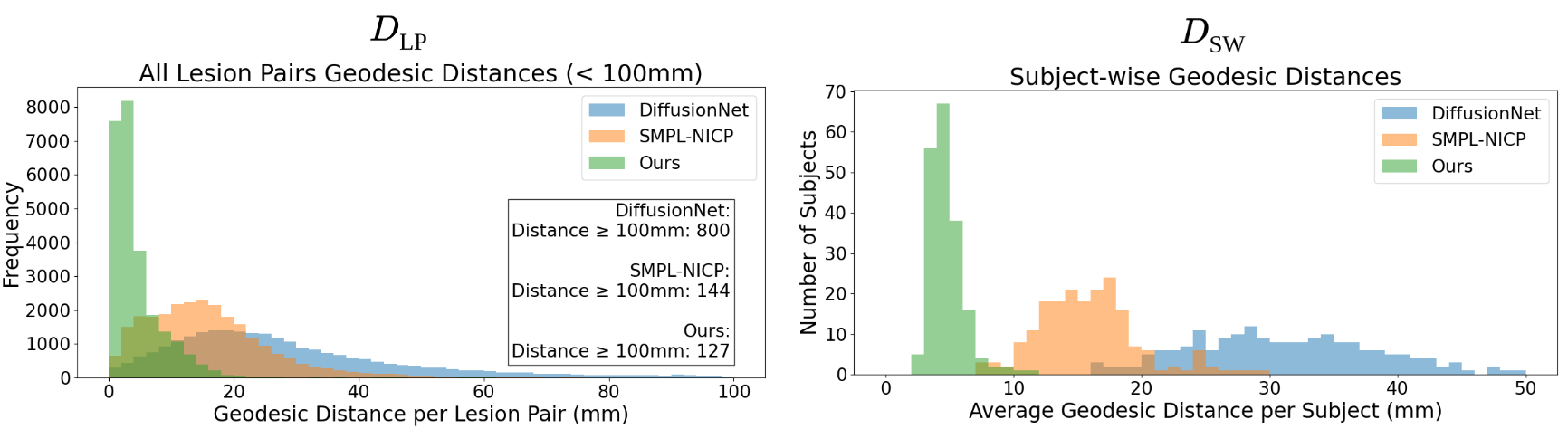}
    \end{center}
   \caption[example] 
    { \label{fig:eval phi all}
(a) The distribution of all lesion pair geodesic distances. (b) The distribution of the subject-wise geodesic distance between lesion pairs.} 
\end{figure}

\begin{figure}[]
   \begin{center}
   \includegraphics[width=\textwidth]{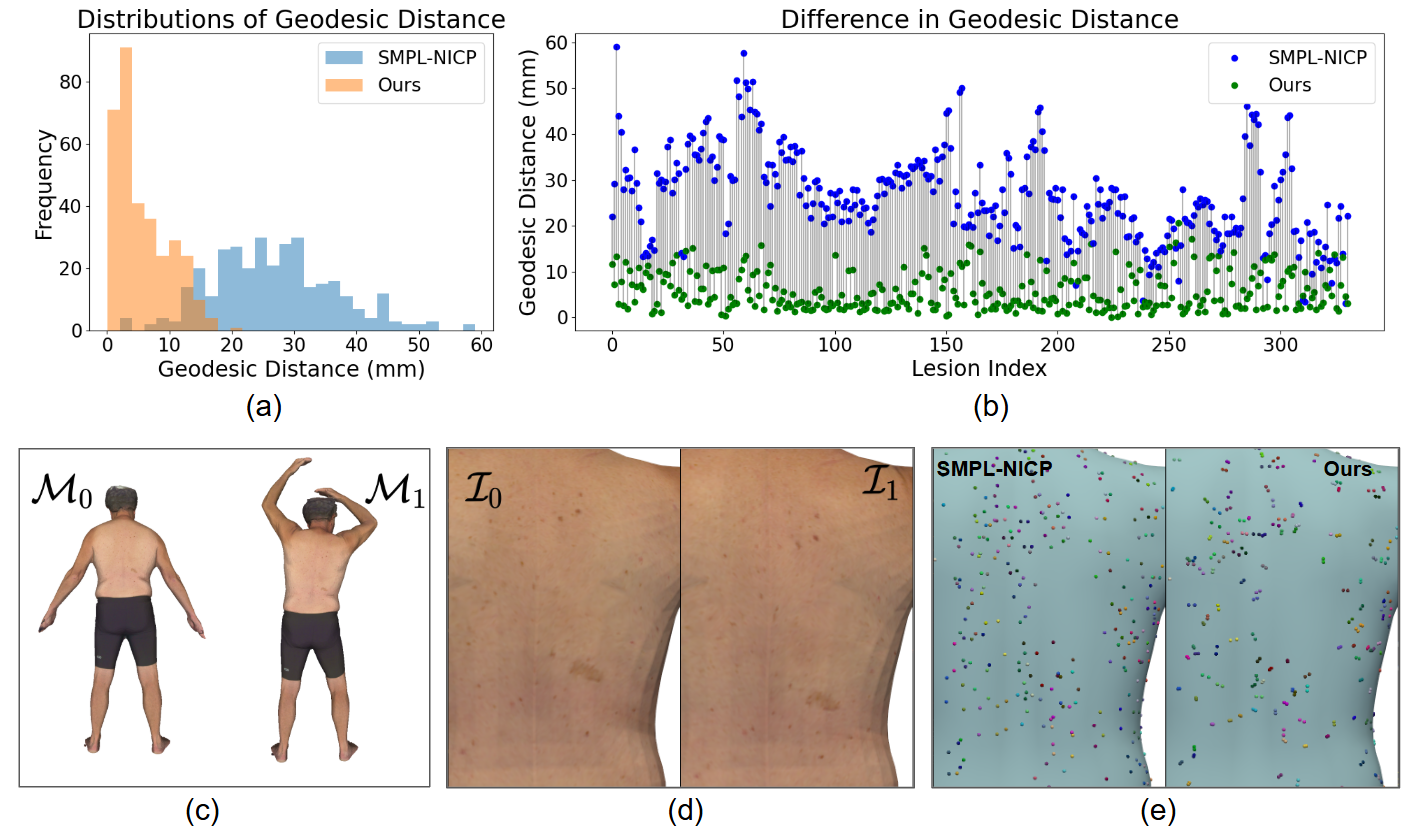}
	\end{center}
   \caption[example] 
    { \label{fig:eval phi individual}
Comparison of correspondence maps for an individual subject (subject-162) between SMPL-NICP and the proposed method. (a) shows the distribution of the geodesic distance for all the lesion pairs on the subject. (b) shows the difference in geodesic distance of each lesion pair between SMPL-NICP and the proposed method. (c) visualizes the source and the target textured meshes. (d) visualizes the source and the target texture signals brought onto the template mesh. (e) visualizes the source and the target lesions mapped to the template mesh using SMPL-NICP and the proposed method. Lesions in correspondences are in the same color.} 
\end{figure}

Fig.~\ref{fig:eval phi all} (a) shows the distribution of the geodesic distance between all the lesion pairs.
Fig.~\ref{fig:eval phi all} (b) shows the distribution of the subject-wise geodesic distance between lesion pairs.
We observe that the proposed method effectively aligns lesion pairs closer and reduces the long-tail distribution as well.
To further investigate the improvement of flow field refinement, we compare correspondence maps established by the proposed method and SMPL-NICP for an individual subject. Fig.~\ref{fig:eval phi individual} (a) shows the distribution of the geodesic distances for all the lesion pairs on the subject. In Fig.~\ref{fig:eval phi individual} (b), we observe that the flow field refinement is also effective in aligning lesion pairs that are originally far from each other (e.g. with a geodesic distance of more than 30 mm). Fig.~\ref{fig:eval phi individual} (c) visualizes the source and the target mesh of the subject, demonstrating challenges of non-isometric deformation due to soft tissue and differing poses. Fig.~\ref{fig:eval phi individual} (d) visualizes the texture signals transferred to the template mesh. In Fig.~\ref{fig:eval phi individual} (e), we show the source and the target lesions mapped onto the template mesh with (ours) and without (SMPL-NICP) flow field refinement. We observe that after the refinement source and target lesions are mapped more closely together, facilitating the task of lesion assignment.

\begin{table}[ht]
\caption{Comparison of the accuracy of the established correspondence maps. $D_{\text{LP}}$ denotes the geodesic distance across all lesion pairs. $D_{\text{SW}}$ represents the geodesic distance between lesion pairs across all the subjects. The standard deviation is shown in brackets.} 
\label{tab:phi comparison}
\centering
\small
\begin{tabular}{|l|l|l|l|}
\hline
\rule[-1ex]{0pt}{3.5ex} & DiffusionNet \cite{sharp2022diffusionnet} & SMPL-NICP \cite{marin24nicp}  & Ours \\
\hline
$D_{\text{LP}}$ (mm) & 31.6 (28.5) & 16.2 (12.2) & 4.9 (8.4) \\
\hline
$D_{\text{SW}}$ (mm) & 32 (8.8) & 15.9 (4) & 4.8 (1.4) \\
\hline
Success rate (\%) & 12 (7) & 30.9 (13.9) & 89.9 (4.7) \\
\hline
\end{tabular}
\end{table}

\subsection{Evaluation of Lesion Assignment}
\label{ss:evaluation pi}
We compare the proposed framework to existing lesion matching methods from Zhao \textit{et al.} \cite{zhao2022skin3d} and Ahmedt-Aristizabal \textit{et al.} \cite{ahmedt2023monitoring}. We note that the two methods essentially rely on the anatomical position of the source lesions and the target lesions that are mapped in the template mesh while differing in two ways. First, Ahmedt-Aristizabal \textit{et al.} \cite{ahmedt2023monitoring} use linear assignment, whereas Zhao \textit{et al.} \cite{zhao2022skin3d} resort to quadratic assignment using the preservation of geodesic distance between lesion pairs.
Second, Ahmedt-Aristizabal \textit{et al.} \cite{ahmedt2023monitoring} selects LoopReg \cite{bhatnagar2020loopreg} for template registration, while Zhao \textit{et al.} \cite{zhao2022skin3d} selects 3D-CODED \cite{groueix20183d}. To eliminate the effect coming from different registering methods, we re-implement their methods using SMPL-NICP \cite{marin24nicp} for template registration, similar to what is used in our framework. 

For each subject, we calculate the matching accuracy as the number of correctly matched lesions over the number of annotated lesion pairs. Table~\ref{tab:pi comparison} shows the comparison of the matching accuracy evaluated with different subsets of the dataset. The matching accuracy of the proposed method is the highest for the entire dataset and the ``numerous-lesions'' subset. In particular, when only evaluating subjects within the ``numerous-lesions'' subset, we observe that the improvement from our method is more pronounced. In this subset, our method achieves a 98.2\% ($\sigma=1.5\%$) matching accuracy over 35 subjects totaling 9772 lesion pairs, as compared to 92.2\% ($\sigma=16.9\%$) and 96.0\% ($\sigma=5.7\%$) for the methods of Zhao \textit{et al.} \cite{zhao2022skin3d} and Ahmedt-Aristizabal \textit{et al.} \cite{ahmedt2023monitoring}, respectively. Therefore, the proposed method effectively pairs up skin lesions for subjects with numerous lesions, the most important benefit of using TBP for skin cancer. The matching accuracy for individual subjects within the ``numerous-lesions'' subset is shown in the supplement. We remark that the reported matching accuracy from Zhao \textit{et al.} \cite{zhao2022skin3d} is 83\% ($\sigma=38\%$) conducted on 10 subjects and totaling around 200 lesion pairs. We also observe that the proposed method is more robust to differences in topology between source and target meshes. A subject with changes in topology between the two scans and the results can be found in the supplement. 

For the ``challenging-pose'' subset, the approach from Zhao \textit{et al.} \cite{zhao2022skin3d} gives the highest matching accuracy, followed by our method. We note that the template-based correspondence map fails when the poses are challenging, either unseen in the training dataset (in SMPL-NICP \cite{marin24nicp}) or far from the template ``T'' pose. In this case, comparing the methods from Zhao \textit{et al.} \cite{zhao2022skin3d} and Ahmedt-Aristizabal \textit{et al.} \cite{ahmedt2023monitoring}, the preservation of the geodesic distance between pairs of lesions helps to improve the matching accuracy beyond linear assignment.

\begin{table}[ht]
\caption{Comparison of the matching accuracy (\%) across all the subjects within different subsets of the dataset. The standard deviation is shown in brackets. Recall that the ``entire'' subset has 192 subjects, excluding 6 subjects in the ``challenging-pose'' subset. There are 35 subjects in the ``numerous-lesions'' subset.} 
\label{tab:pi comparison}
\centering
\small
\begin{tabular}{|l|l|l|l|} 
\hline
\rule[-1ex]{0pt}{3.5ex} & Zhao \textit{et al.} \cite{zhao2022skin3d} & Ahmedt-Aristizabal \textit{et al.} \cite{ahmedt2023monitoring} & Ours \\
\hline
\rule[-1ex]{0pt}{3.5ex}  entire  & 95.9 (11.6) & 98.3 (3.6) & 98.9 (1.7)\\
\hline
\rule[-1ex]{0pt}{3.5ex}  numerous-lesions  & 92.2 (16.9) & 96.0 (5.7) & 98.2 (1.5)\\
\hline
\rule[-1ex]{0pt}{3.5ex}  challenging-pose  & 87.0 (8.8)  & 72.2 (8.7) & 80.4 (9.6)\\
\hline
\end{tabular}
\end{table}

\subsection{Robustness to Noise in Lesion Detection}
We evaluate the robustness of the proposed framework to noisy lesion detection by independently taking out $p$\% of lesions in the source and target, and then pairing up lesions. The evaluation is performed on the ``numerous-lesions'' subset. We compare the proposed framework to the approaches from Zhao \textit{et al.} \cite{zhao2022skin3d} and Ahmedt-Aristizabal \textit{et al.} \cite{ahmedt2023monitoring}.

We compute precision, recall, and F1 scores between the source and target lesions for each subject and then report the average and standard deviation across subjects. The F1 score is the harmonic mean of precision and recall. Given predicted matrix $\pi_{pred}$ containing $k_{pred}$ matches, and ground truth matrix $\pi_{gt}$ containing $k_{gt}$ matches, denote $\odot$ as element-wise product, we have:
\begin{align}
    \text{precision} &= \sum(\pi_{pred} \odot \pi_{gt})/k_{pred}, \label{eq:precision} \\
    \text{recall} &= \sum(\pi_{pred} \odot \pi_{gt})/k_{gt}, \label{eq:recall} \\
    \text{F1} &= 2\cdot \frac{\text{precision} \cdot \text{recall}}{\text{precision} + \text{recall}}. \label{eq:f1}
\end{align}

\begin{figure}[]
   \begin{center}
   \includegraphics[width=\textwidth]{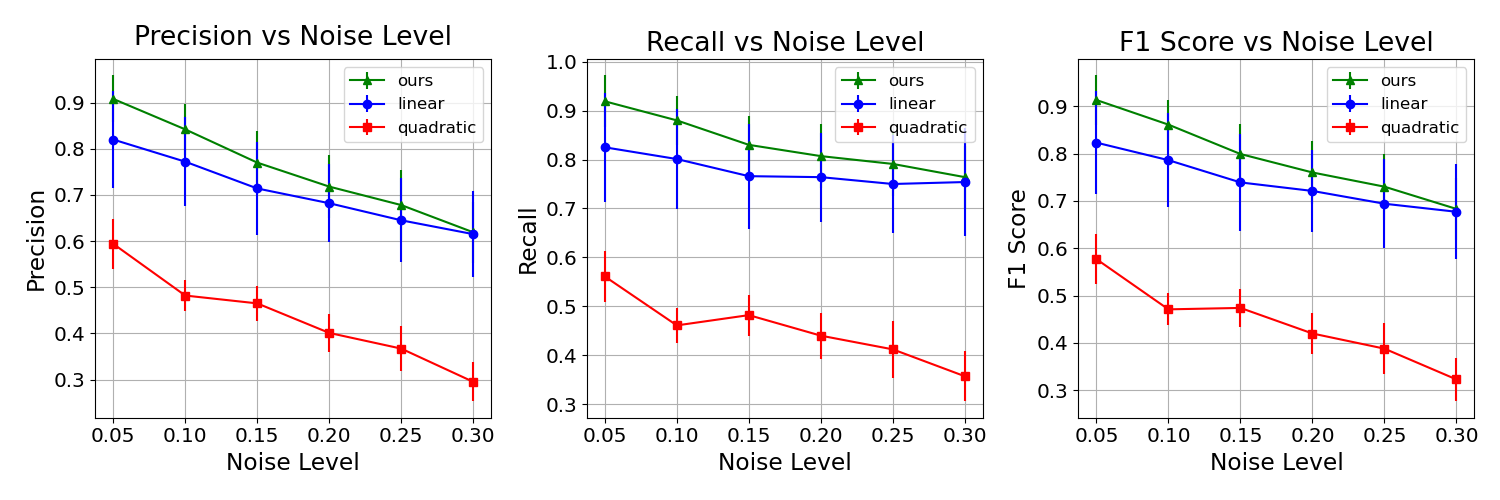}
	\end{center}
   \caption[example] 
    { \label{fig:exp noisy}
Comparison of the precision, recall, and F1 scores under different noise levels. ``linear'' and ``quadratic'' represent the methods of Ahmedt-Aristizabal \textit{et al.} \cite{ahmedt2023monitoring} and Zhao \textit{et al.} \cite{zhao2022skin3d}, respectively. The noise level is the percentage of lesions independently taken out from the source and target lesions, ranging from 5\% to 30\% every 5\%.
} 
\end{figure}

Fig.~\ref{fig:exp noisy} compares the precision, recall, and F1 scores for the three methods under different noise levels. Compared to the baseline methods, the proposed method is consistently more robust to errors in lesion detection for a noise level smaller than 25\%. We note that the reported recall for lesion detection by Zhao \textit{et al.} \cite{zhao2022skin3d} ranges from 78\% to 96\%, validating that the noise range in our experiment is practical.
Furthermore, while the preservation of geodesic distance helps the lesion matching when correspondence maps fail to align lesions (in \S\ref{ss:evaluation pi}), it is notoriously sensitive to noise, which is corroborated in our findings. 

\subsection{Ablation Study}
\subsubsection{Initial correspondence map}
Conceptually, the proposed flow field refinement can be applied to any shape correspondence method that maps the source and target mesh to a template mesh. However, the level of improvement might be different, depending on the consistency of the source and target signals constructed on the template mesh.
Therefore, we compare how the flow field refinement works on different initial correspondence maps. We evaluate the average geodesic distance across all the annotated lesion pairs ($D_{LP}$), using DiffusionNet \cite{sharp2022diffusionnet} and SMPL-NICP \cite{marin24nicp} for initial correspondence maps. The $D_{LP}$ for DiffusionNet and SMPL-NICP are 31.6 mm ($\sigma=28.5$) and 16.2 mm ($\sigma=12.2$) initially, and 27.4 mm ($\sigma=17.9$) and 4.9 mm ($\sigma=8.4$) after refinement, respectively. We show that the proposed flow field refinement effectively improves the correspondence map from both methods. Remarkably, the improvement for a more precise initial correspondence map is more signnificant -- an improvement of 11.3 mm for SMPL-NICP as compared to an improvement of 4.2 mm for DiffusionNet. Since SMPL-NICP gives more consistent source and target signals to be aligned, in line with the assumptions of brightness constancy and little motion \cite{lucas1981iterative,tomasi1991detection}, it benefits more from flow field refinement.

\subsubsection{Flow field refinement}
\paragraph{Usage of signals}
To investigate the effectiveness of different signals to refine the correspondence maps, we compare $D_{LP}$ by using only the texture signal, the lesion signal, and a combination of the two. On the ``numerous-lesion'' subset, we observe that using the texture signal gives 14.4 mm ($\sigma=12.7$), while there is no significant difference between using only the lesion signal compared to using a combination of the two, with both giving 4.8 mm ($\sigma=6.9$). 
Overall, the signals used in our method refine the correspondence maps while being robust to inconsistent source and target texture coming from scanning artifacts. 
\paragraph{Usage of small magnitude regularization}
To respect the initial correspondence map and be robust to inconsistency in signals due to noise (e.g. undetected lesions or texture that does not agree in the source and target), we add a magnitude regularization to enforce the vector field to be small. On the ``numerous-lesion'' subset, we observe that the $D_{LP}$ is 15.84 mm ($\sigma=11.5$) without the regularization and 4.8 mm ($\sigma=6.9$) with the regularization.

\section{Discussion}

\subsection{Choice of Barycentric Coordinate Representation}
The choice of the barycentric coordinate representation for lesions (and correspondence maps) tackles the limitation of inaccurate and resolution-dependent matching present in previous works \cite{zhao2022skin3d,ahmedt2023monitoring,huang2023skin}.  
As a comparison, on the entire dataset, the average matching accuracy across all the subjects using the vertex representation is 82.9\% ($\sigma=11.2$), as compared to 98.3\% ($\sigma=3.6$) evaluated with our re-implementation of approach from Ahmedt-Aristizabal \textit{et al.} \cite{ahmedt2023monitoring}. As expected, we observe a consistent decrease in matching accuracy as the lesion count increases due to inaccurate representation. A detailed comparison of the average matching accuracy for different numbers of lesions per subject can be found in the supplement. 
In addition, the selected representation is robust to the mesh resolution. Our evaluation is conducted on the low-resolution mesh from the 3DBodyTex dataset \cite{saint20183dbodytex,saint2019bodyfitr} with 10K vertices on average, whereas both Zhao \textit{et al.} \cite{zhao2022skin3d} and Haung \text{et al.} \cite{huang2023skin} use high-resolution mesh with 300K vertices from the same dataset.

\subsection{Computational cost}
Our method achieves state-of-the-art matching accuracy without sacrificing efficiency. For each subject, our framework takes around 7 minutes, including the following components: 1) Coarse correspondence map (4 minutes). 2) Signal construction on template mesh (20 seconds). 3) Surface optical flow (150 seconds). 4) Lesion assignment (1 second). In particular, for the ``numerous-lesion'' subset, our method performs significantly better than the approach from Ahmedt-Aristizabal \textit{et al.} \cite{ahmedt2023monitoring} with an acceptable computation overhead of 3 minutes. On the other hand, the quadratic assignment used by Zhao \textit{et al.} \cite{zhao2022skin3d} is NP-hard.

\subsection{Evaluation on Subjects with Dark Skin Tones}
Our dataset includes a variety of skin tones, notably including 5 subjects with dark skin tones. We observe that the number of skin lesions is relatively small for those subjects, with 26.6 ($\sigma=7.7$) lesions on average. Our method successfully pairs up all the lesions on the 5 subjects. Example results can be found in the supplement.

\subsection{Limitations}
\begin{figure}[]
   \begin{center}
   \includegraphics[width=\textwidth]{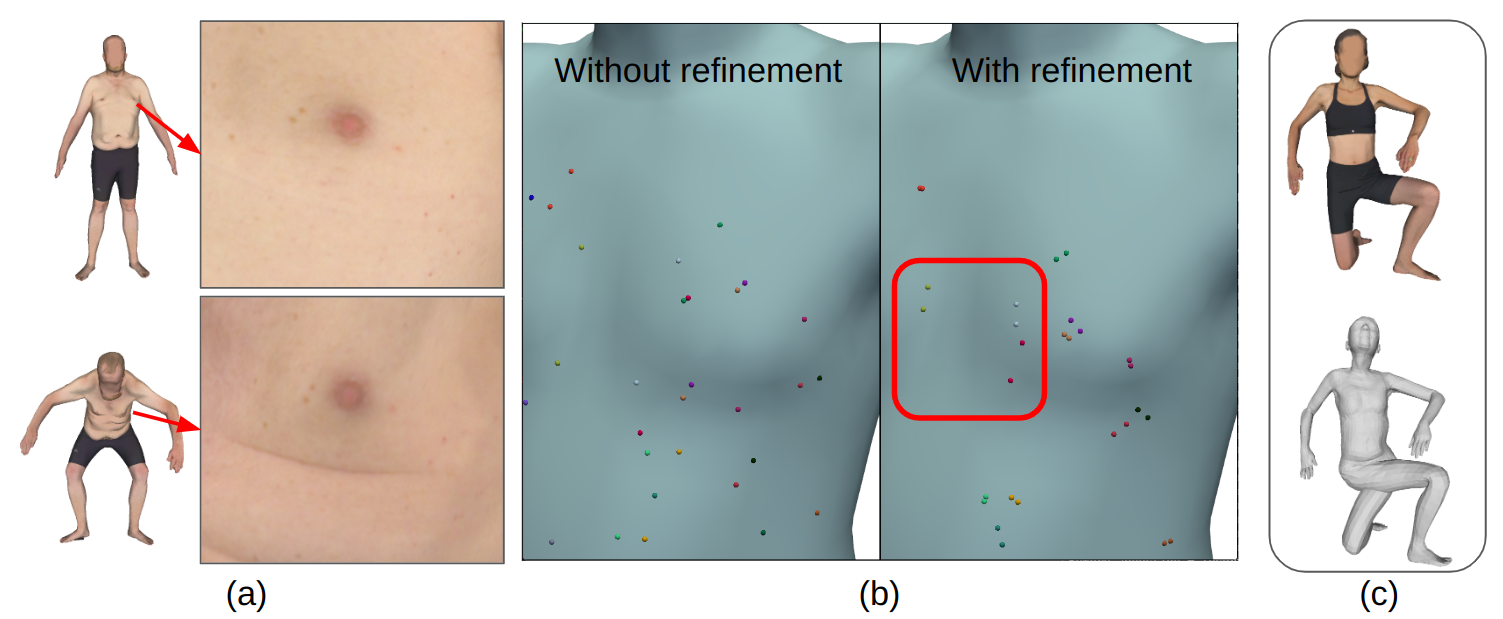}
	\end{center}
   \caption[example] 
    { \label{fig:failure}
Examples of failure cases. (a) visualizes a subject undergoing large non-isometric deformation. (b) illustrates the source and the target lesions imaged on the template mesh for lesions shown in (a). In the red box, we highlight lesion pairs with a geodesic distance of more than 50 mm. (c) shows a failure in template registration.
} 
\end{figure}

The proposed framework has several limitations.
First, our method fails to accurately map the source and the target lesions on the template mesh within the 10-mm criterion if a large non-isometric deformation exists. Fig.~\ref{fig:failure} (a) visualizes a subject with large non-isometric deformation around the chest and the belly area. Fig.~\ref{fig:failure} (b) demonstrates the source and the target lesions imaged on the template mesh. Although the proposed vector-field-based refinement brings closer lesions in correspondence, the lesion pairs in the red box are still far from each other, with a geodesic distance of more than 50 mm. However, these lesions are correctly paired up in our lesion assignment step. Second, the proposed framework fails for some challenging poses when the template mesh is incorrectly registered to the input mesh. For example, Fig.~\ref{fig:failure} (c) shows an example where template-based registration fails. The registered template mesh flips the left and the right legs. As a result, our method cannot match lesions successfully for lesions on those incorrectly registered body regions. However, some challenging poses may not be common in clinical settings, especially considering that patients have to maintain the pose during the scan.
Last, the proposed dataset exhibits limited annotations of distal extremities, such as lesions on fingers and toes. In particular, due to complex morphology, these anatomical regions are challenging for digital imaging. Consequently, the proposed framework may demonstrate reduced performance and reliability for lesions in these regions.


\section{Conclusions}
In this paper, we propose a framework to match lesions in the context of full body using 3D textured meshes while providing locations for unmatchable lesions. We propose a skin lesion tracking dataset with 25K lesion pairs over 198 subjects. As far as we know, we are the first to release a dataset for skin lesion tracking in 3D TBP at this scale. 
We show that the proposed method effectively refines correspondence maps to align lesion pairs, achieving a success rate of 89.9\% at 10-mm criterion. The proposed framework accomplishes state-of-the-art matching accuracy, an accuracy of 98\% for 35 subjects with more than 200 lesions on the body. Furthermore, our method is validated to be more robust to inconsistent texture between the source and the target meshes and less sensitive to errors in lesion detection.

In the future, we would like to extend the framework for more than two TBP scans. With more than two TBP scans, some of the false positives and false negatives may potentially be resolved by evaluating the consistency of a lesion's life cycle \cite{di2023graph}. Moreover, the method needs to be evaluated on longitudinal data with a longer duration that may include more significant changes in skin conditions.

\subsubsection{Acknowledgments}       

The research was in part supported by the Intramural Research Program (IRP) of the NIH/NICHD, Phase I of NSF STTR grant 2127051, Phase II of NSF STTR grant 2335086, and Phase I NIH/NIBIB STTR grant R41EB032304. We thank Ryan Whittaker for the help with annotating skin lesion pairs.
 
%
%
%
\bibliographystyle{splncs04}
\bibliography{references}
\end{document}